\pdfoutput=1
\documentclass[11pt]{article}
\usepackage{acl}
\usepackage{times}
\usepackage{latexsym}
\usepackage[T1]{fontenc}
\usepackage[utf8]{inputenc}
\usepackage{microtype}
\usepackage{enumitem}
\usepackage{inconsolata}
\usepackage{amsmath} 
\usepackage{amssymb}
\usepackage{booktabs}  
\usepackage{xspace}
\usepackage{tabularx}  
\usepackage{geometry}  
\usepackage{placeins}
\usepackage{stfloats}
\usepackage{array} 
\usepackage{textcomp}
\usepackage{graphicx}
\usepackage[table]{xcolor}
\usepackage{bm}
\usepackage{algorithm}
\usepackage{algpseudocode}
\usepackage{listings}

\newcolumntype{Y}{>{\raggedright\arraybackslash}X} 

\newcommand{\llm}{Xmodel-2.5\xspace}

\title{\llm: 1.3B Data-Efficient Reasoning SLM}

\author{
Yang Liu
\hspace{0.8em}Xiaolong Zhong
\hspace{0.8em}Ling Jiang \\ \\
Xiaoduo AI Lab \\
\texttt{foamliu@yeah.net}
}

\begin{document}
\maketitle
\begin{abstract}
Large language models deliver strong reasoning and tool-use skills, yet their computational demands make them impractical for edge or cost-sensitive deployments.
We present \textbf{Xmodel-2.5}, a 1.3-billion-parameter small language model designed as a \emph{drop-in agent core}.
Training with maximal-update parameterization ($\mu$P) allows hyper-parameters tuned on a 20M-parameter proxy to transfer directly to the full model, even under the parameter-tied \emph{tie-word-embedding} architecture.
A 1.4T-token Warmup--Stable--Decay curriculum is used, and we further show that \textbf{switching from AdamW to Muon during the decay phase} improves the 13-task reasoning average by 4.58\,\% while keeping every other hyper-parameter fixed, verifying that early AdamW stability can be paired with late Muon sharpening for better downstream performance.
FP8-mixed-precision training balances accuracy and throughput.
All checkpoints, recipes, and evaluation code are released under the Apache-2.0 license.\footnote{\url{https://huggingface.co/XiaoduoAILab/Xmodel-2.5} and \url{https://huggingface.co/XiaoduoAILab/Xmodel-2.5-history} (training checkpoints).} 
Training code and evaluation harness: {\url{https://github.com/XiaoduoAILab/Xmodel-2.5}}.
\end{abstract}
\section{Introduction}

Large language models (LLMs) have demonstrated remarkable reasoning, planning, and tool-use capabilities, yet their deployment as autonomous agents remains prohibitive for resource-constrained environments.
State-of-the-art agent backbones typically exceed 7--13\,B parameters, demanding high-end accelerators and large memory footprints incompatible with edge or cost-sensitive scenarios.

Recent small language models (SLMs, $<$\,2\,B) match LLMs on single-turn benchmarks such as GSM8K or MMLU, but still fall short in \emph{complex multi-step reasoning}—the core skill required for tool invocation, long-horizon planning, and robust error recovery.
Boosting this capability within the 1--2\,B regime is the central goal of our work.

\paragraph{Xmodel-2.5}
We present Xmodel-2.5, a 1.3\,B-parameter decoder-only model that retains Xmodel-2's two-stage pre-training recipe and maximal-update parameterization ($\mu$P), while introducing four targeted upgrades:

\begin{enumerate}
  \item We extended \textbf{Megatron-LM} with complete $\mu$P support, modifying its parameterization, attention scaling, and residual connections. The implementation was validated to preserve $\mu$P dynamics, enabling reliable hyperparameter transfer.

  \item \textbf{Tokenizer.} Adopted the 129\,k-token DeepSeek-v3 tokenizer (vs.\ Xmodel-2's 65\,k-token Unigram), improving compression and decoding speed.

  \item \textbf{Numeric precision.} Switched from BF16 to FP8-mixed precision, raising training throughput by $\approx 30\%$ with no observable degradation in pilot experiments.

  \item \textbf{Optimizer schedule.} Switched from AdamW to \textbf{Muon} during the decay phase, improving the 13-task reasoning average by 4.58\% while keeping all other hyper-parameters fixed.

\end{enumerate}

We hope Xmodel-2.5 serves as a \emph{minimal yet strong} baseline for lightweight agents with enhanced complex-reasoning capabilities.
\section{Background \& Related Work}

\subsection{Small Language Models for Reasoning}
Parameter-efficient SLMs ($<2$\,B) have recently closed the gap with larger counterparts on mathematical and commonsense reasoning.
MiniCPM~\citep{minicpm4} and DCLM-1B~\citep{dclm2024} adopt code-enriched corpora and cosine or WSD schedules to surpass 35\% on GSM8K.
\citet{phi35} further emphasises textbook-style synthetic data.
These works, however, primarily target single-turn question answering; systematic evaluation of multi-step \emph{agentic} behaviours remains under-explored.

\subsection{Hyper-Parameter Transfer with Maximal-Update Parameterisation}
$\mu$P~\citep{yang2022tensor,yang2023tensor} preserves training dynamics across widths, enabling hyper-parameter transfer from ``toy'' to full-scale models.
Original derivations assume SGD; recent work integrates Adam~\citep{adam2017} but reports instability below 2\,B parameters.

\subsection{Efficient Training of Lightweight Models}
FP8 mixed-precision~\citep{fp8nvidia2022} and fused attention kernels reduce memory and energy, yet have not been jointly studied with $\mu$P optimisers.
WSD learning-rate schedules~\citep{minicpm2024} improve late-stage performance by decoupling the annealing phase from token count; we extend WSD with domain-weighted data mixing during decay, an ablation absent in prior literature.
\section{Methodology}\label{sec:method}

We scale Xmodel-2 to 1.3\,B parameters while retaining its deployment-friendly, deep-and-thin decoder-only skeleton.
The section below details three design pillars:
(i) architecture-level $\mu$P compatibility (\S\ref{sec:arch}),
(ii) a three-phase Warmup--Stable--Decay (WSD) curriculum (\S\ref{sec:wsd}), and
(iii) FP8 mixed-precision acceleration (\S\ref{sec:fp8-detail}).

\subsection{Model Architecture}\label{sec:arch}
Xmodel-2.5 keeps the deep-and-thin decoder-only design of Xmodel-2, with the configuration in Table~\ref{tab:config}.
To preserve maximal-update dynamics across widths we apply the $\mu$P attention-logits scaling $1/d_{\text{head}}$; all other components (Pre-RMSNorm, SwiGLU) are inherited without modification.
While Xmodel-2 was trained with Flash-Attention v2, Xmodel-2.5 uses the CuDNN backend provided by Megatron-LM (no \texttt{--use-flash-attn} specified).

\begin{table}[htbp]
\centering
\small
\begin{tabular}{lc}
\toprule
\textbf{Hyper-parameter} & \textbf{Value} \\
\midrule
Hidden size & 1536 \\
Intermediate size & 3840 \\
Num of transformer layers & 48 \\
Attention heads (Q) & 24 \\
KV heads (GQA) & 8 \\
Sequence length & 3712 \\
Max position embeddings & 131072 \\
RoPE base & 500000 \\
\bottomrule
\end{tabular}
\caption{Model configuration. {\footnotesize Long-context support via 131\,K position embeddings and RoPE base 500\,K.}}
\label{tab:config}
\end{table}

\subsection{Three-Phase WSD Pre-Training}\label{sec:wsd}

The 560\,k-step Warmup--Stable--Decay (WSD) schedule consumes 1.4\,T tokens:
\begin{enumerate}
  \item \textbf{Warmup (W)}: 2\,k steps, LR linearly rises to $1.67\!\times\!10^{-3}$ (hidden) and $0.01$ (embeddings).
  \item \textbf{Stable (S1)}: 270\,k steps, batch size 3\,712$\times$480$\approx$1.78\,M tokens.
  \item \textbf{Stable (S2)}: 260\,k steps, batch size 3\,712$\times$960$\approx$3.56\,M tokens.
  \item \textbf{Decay (D)}: 20\,k steps (exponential decay), batch size $\approx$3.56\,M tokens, 66.9\,\% high-quality SFT data blended.
  \item \textbf{Long-Context Reasoning (LCR)}: 10\,k additional steps, \emph{same} 66.9\,\% SFT ratio, batch size 16\,384$\times$240$\approx$3.93\,M tokens, context length 16\,k.
\end{enumerate}

\paragraph{Optimizer ablation: AdamW $\rightarrow$ Muon.}
During the final 20\,k decay steps we perform a controlled optimizer switch: AdamW is replaced by Muon~\citep{jordan2024muon} while \emph{all other hyper-parameters remain frozen}, including the learning-rate schedule, batch size, gradient-clip, weight-decay, and the 66.9\,\% SFT data blend.
Under this single-variable change, the 13-task reasoning average rises by 4.58\,pp \emph{absolute} over an AdamW-only baseline, confirming that the gain is attributable to Muon and not to confounding factors such as data re-weighting or schedule re-tuning.
This result supports the view that early-phase stability (AdamW) can be complementarily combined with late-phase sharpness (Muon) to improve downstream performance without extra data or re-training.

\begin{figure*}[htbp]
    \centering
    \includegraphics[width=0.95\textwidth]{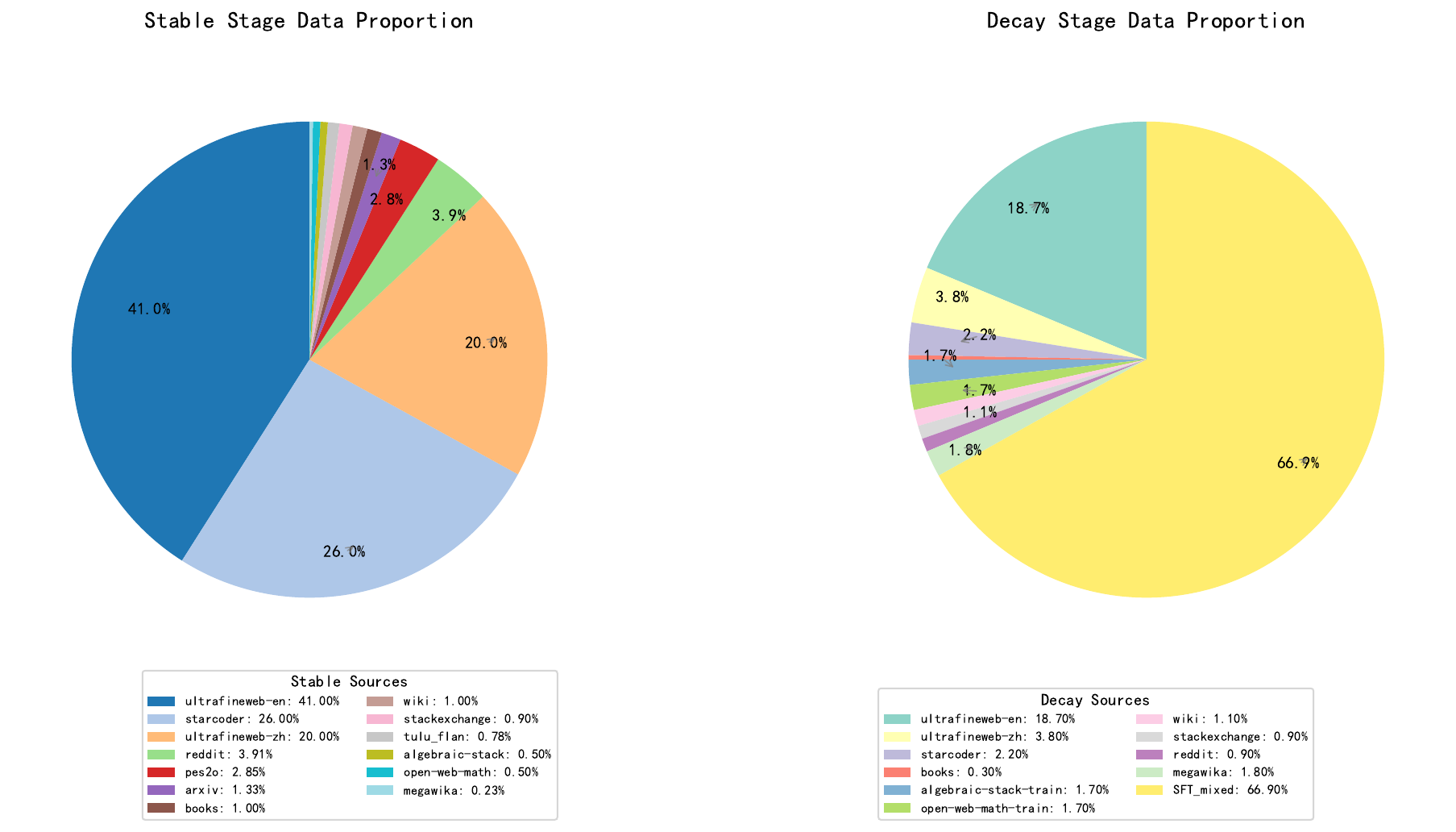}
    \caption{Data composition in the (a) Stable and (b) Decay phases of WSD LR scheduling. The Stable phase emphasizes broad pre-training data diversity, while the Decay phase focuses on high-quality instructional and SFT data to refine model capabilities.}
    \label{fig:data_composition}
\end{figure*}

\subsection{FP8 Hybrid Format \& Kernel Implementation}
\label{sec:fp8-detail}

We adopt the FP8 hybrid format implemented in NVIDIA Transformer Engine~(TE)~\cite{nvidia-transformer-engine}:
\begin{itemize}[leftmargin=*,nosep]
  \item \textbf{Forward}: E4M3 (1-4-3 exponent-mantissa) for activations;
  \item \textbf{Backward}: E5M2 (1-5-2) for gradients;
  \item \textbf{Master-weights}: kept in \texttt{bfloat16} to avoid under-flow.
\end{itemize}
TE delayed-scaling hyper-parameters: \texttt{amax-history-len}=128, \texttt{amax-compute-algo}=\texttt{max}.  
All FP8 kernels are invoked through Megatron-LM's \texttt{--transformer-impl transformer\_engine} flag;  
the corresponding GEMM, Layer-Norm and GeLU CUDA kernels are automatically selected without source-code modification. 
\section{Experimental Settings}

\subsection{Baselines}
We compare Xmodel-2.5 with eight publicly released decoder-only models in the 1--2\,B range:

\begin{itemize}
  \item \textbf{Qwen3-1.7B}~\citep{yang2025qwen3} — updated Qwen series with enhanced code and math data;
  \item \textbf{MiniCPM-1B}~\citep{minicpm2024} — SLM trained with WSD and domain-weighted data;
  \item \textbf{InternLM2.5-1.8B}~\citep{cai2024internlm2} — upgraded InternLM with improved reasoning;
  \item \textbf{Llama-3.2-1B}~\citep{llama3} — Meta's lightweight Llama-3 variant;
  \item \textbf{Gemma-3\_1B}~\citep{gemmateam2025gemma3} — Gemini-2.0-based SLM with alternating local-global attention for 32 k long context.;
  \item \textbf{SmolLM2-1.7B}~\citep{allal2025smollm2} — high-quality corpus focused common-sense model.
  \item \textbf{Xmode2-1.2B}~\citep{qun2024xmodel2} — compact SLM trained with WSD scheduling and math-enriched corpora for strong reasoning.
\end{itemize}

All baseline results are reproduced with our unified evaluation pipeline using the Language Model Evaluation Harness \citep{eval-harness} to ensure fair comparison.

\subsection{Tasks and Metrics}
We evaluate on 13 datasets covering commonsense, symbolic and multilingual reasoning.

\paragraph{Commonsense} ARC-Challenge \citep{clark2018think} (25-shot), ARC-Easy \citep{clark2018think} (25-shot), PIQA \citep{Bisk2019PIQARA} (0-shot), HellaSwag \citep{zellers2019hellaswag} (10-shot), WinoGrande \citep{Sakaguchi2021WinoGrande} (5-shot).

\paragraph{Symbolic Reasoning} BBH \citep{Suzgun2022ChallengingBT} (3-shot), MMLU \citep{hendryckstest2021} (5-shot).

\paragraph{Mathematical \& Code} GSM8k \citep{cobbe2021gsm8k} exact-match with flexible-extract (5-shot), MATH \citep{hendrycks2021math} (4-shot) verified by math\_verify, HumanEval \citep{chen2021humaneval}(0-shot), MBPP \citep{austin2021mbpp}(3-shot).

\paragraph{Chinese Understanding} C-Eval \citep{huang2023ceval} (5-shot), CMMLU \citep{li2024cmmlu} (5-shot).

Metrics: accuracy for all sets; pass@1 for HumanEval/MBPP; exact-match for GSM8k/MATH.

\section{Results}

\subsection{Main Reasoning Results}
Table~\ref{tab:full-results} summarizes zero-shot and few-shot performance on 13 commonsense and mathematical reasoning benchmarks. Despite being 25\% smaller (1.3\,B vs.\ 1.7\,B parameters) and trained with 96 \% fewer tokens (1.4\,T vs.\ 36\,T), Xmodel-2.5 closes 71\% of the gap to Qwen3—raising the 1--2\,B average from 50.34\% (Xmodel-2) to \textbf{52.49\%}, +2.15 pp. This result is only 4.47 pp behind Qwen3 (56.96\%), confirming that the WSD schedule extracts superior reasoning efficiency per parameter and per token.

\begin{table*}[ht]
\centering
\footnotesize
\setlength{\tabcolsep}{3.5pt}
\renewcommand{\arraystretch}{1.1}
\begin{tabular}{lccccccccc}
\toprule
\textbf{Dataset (shots)} & \textbf{Qwen3} & \textbf{MiniCPM} & \textbf{InternLM2.5} & \textbf{Llama-3.2} & \textbf{Gemma-3} & \textbf{SmolLM2} & \textbf{Xmodel-2} & \textbf{\llm} \\
\midrule
ARC-Challenge (25)  & 53.07 & 45.31 & 40.96 & 41.47 & 40.36 & 53.50 & 46.16 &  48.89 \\
ARC-Easy (25)       & 79.67 & 75.21 & 71.68 & 71.80 & 69.65 & 80.81 &  76.22 &  76.94 \\
PIQA (0)            & 72.58 & 75.19 & 73.39 & 74.16 & 72.42 & 77.58 &  75.14 &  75.95 \\
HellaSwag (10)      & 60.16 & 67.90 & 59.14 & 59.76 & 55.96 & 73.16 &  64.05 &  67.24 \\
WinoGrande (5)      & 60.54 & 64.48 & 61.88 & 61.25 & 58.48 & 67.88 &  64.25 &  64.64 \\
BBH (3)             & 45.23 & 35.45 & 41.12 & 38.10 & 38.34 & 35.05 &  48.90 &  54.58 \\
MMLU (5)            & 60.24 & 48.75 & 48.87 & 45.44 & 39.87 & 49.99 &  49.98 &  51.81 \\
GSM8k (5)           & 69.29 & 42.00 & 41.77 & 34.80 & 25.70 & 32.37 &  56.56 &  58.98 \\
MATH (4)            & 35.50 & 12.06 & 21.20 & 16.84 & 24.06 & 10.96 &  25.64 &  28.94 \\
HumanEval (0)       & 42.68 & 43.90 & 35.98 & 31.10 & 32.93 & 25.61 &  29.27 &  28.66 \\
MBPP (3)            & 42.60 & 33.40 & 32.80 & 32.60 & 25.00 & 34.40 &  30.80 &  33.00 \\
CMMLU (5)           & 60.47 & 46.52 & 62.03 & 36.99 & 34.11 & 33.60 &  44.29 &  47.16 \\
C-Eval (5)          & 60.40 & 46.21 & 61.59 & 37.00 & 33.51 & 34.55 &  43.16 &  45.54 \\
\midrule
\textbf{Average}    & 56.96 & 48.95 & 50.19 & 44.72 & 42.34 & 46.88 &  50.34 &  \textbf{52.49} \\
\bottomrule
\end{tabular}
\caption{Comprehensive results (\%) on 13 benchmarks. \textbf{Bold} marks best in 1--2\,B range. \llm 1.3B uses flexible-extract for GSM8k and math\_verify for MATH.}
\label{tab:full-results}
\end{table*}

\subsection{Training Loss}
\begin{figure*}[ht]
\centering
\includegraphics[width=0.8\linewidth]{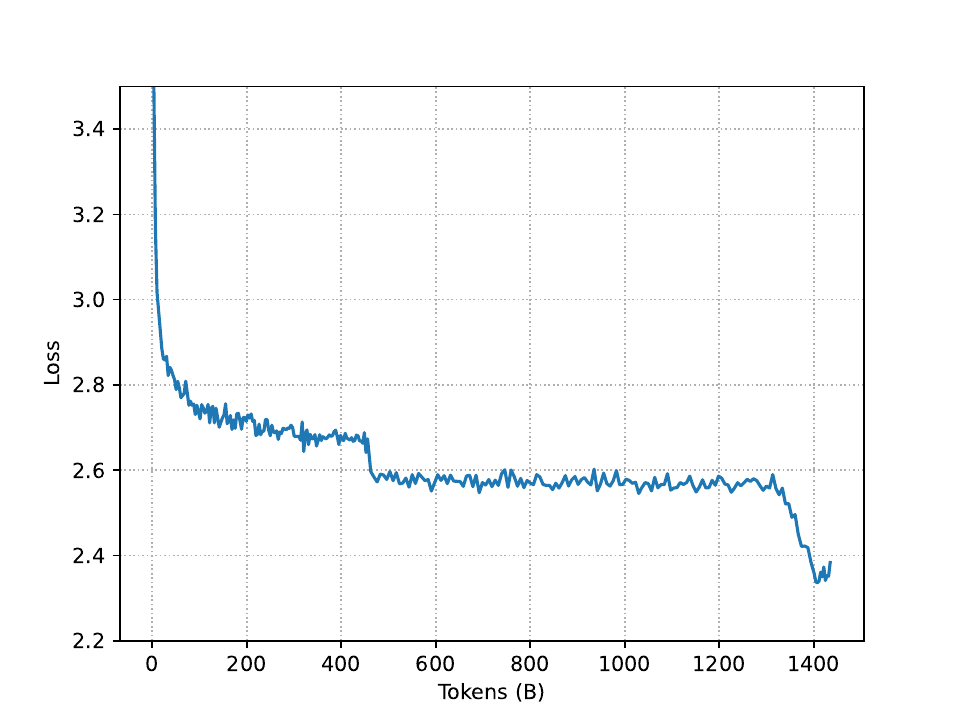}
\caption{Loss curve for \llm 1.3B.}
\label{fig:loss}
\end{figure*}

Figure~\ref{fig:loss} presents the training loss curve on the WikiText-2 dataset \citep{merity2016pointer}. The initial drop corresponds to increasing the batch size from 2M to 4M tokens, which likely replicates the stabilizing effect of a reduced learning rate \citep{smith2018batchsize}. The second drop reflects the impact of the learning rate decay phase. Immediately after decay, we conduct a lightweight long-context adaptation (shaded region in Figure~\ref{fig:loss}): starting from 550k steps, we first expand the context length from 3,712 to 8,192 within 3k steps, and then further to 16,384 within the next 7k steps. As expected, the longer-context exposure produces a small bump on WikiText-2 perplexity; nevertheless, the 13-task average in Table~\ref{tab:full-results} rises from 52.36 to 52.49, confirming that the long-context phase converts the modest perplexity increase into tangible downstream reasoning gains.

\section{Conclusion}
We presented Xmodel-2.5, a 1.3-billion-parameter small language model that achieves the \textbf{second-best} average score among 1--2\,B models on thirteen widely used reasoning benchmarks, trailing only Qwen3 (52.49\,\% vs.\ 56.96\,\%).
Critically, this result is obtained with only 1.4\,T training tokens---25.7$\times$ fewer than Qwen3---demonstrating superior data efficiency.
By extending the Warmup--Stable--Decay schedule with a 10\,k-step long-context adaptation phase, we deliver reliable 16\,k-context reasoning without extra data or hyper-parameter tuning.
Our recipe is simple, compute-friendly, and reproducible, showing that careful pacing and lightweight context stretching can extract more reasoning power from every parameter and every token.
We hope these findings encourage the community to further explore data-efficient pathways toward capable small models.
\bibliography{reference}

\end{document}